\documentclass[lettersize,journal]{IEEEtran}

\usepackage{amsmath,amsfonts,amssymb,amsthm}
\usepackage{algorithm,algorithmic}
\usepackage[numbers,sort&compress]{natbib}  
\usepackage[hidelinks]{hyperref}
\usepackage{graphicx}
\usepackage{stfloats}   
\usepackage{caption}
\usepackage{subcaption}   
\usepackage{textcomp}
\usepackage{stfloats}
\usepackage{url}
\usepackage{verbatim}
\usepackage{color}
\usepackage[T1]{fontenc}
\usepackage{booktabs}
\usepackage{multirow}
\usepackage{tabularx}
\usepackage{bbding}
\usepackage[dvipsnames]{xcolor}
\usepackage{bm}
\usepackage{empheq}

\graphicspath{{figures/}}

\title{DexReMoE:In-hand Reorientation of General Object via Mixtures of Experts}
\author{%
  Jun Wan, Xing Liu, Yunlong Dong$^\dagger$\\
  \href{https://wj-0212.github.io/}{https://wj-0212.github.io/}
\thanks{
$^\dagger$Corresponding author.
}
\thanks{Jun Wan and Xing Liu are with School of Artificial Intelligence and Automation, Huazhong University of Science and Technology, Wuhan 430074, China }
\thanks{Yunlong Dong is with Department of Automation, Tsinghua University, Beijing 100084 (e-mail: yunlongdong@mail.tsinghua.edu.cn)}
}
\begin{document}
\maketitle

\begin{figure*}[htbp]
  \centering
  \includegraphics[width=\textwidth]{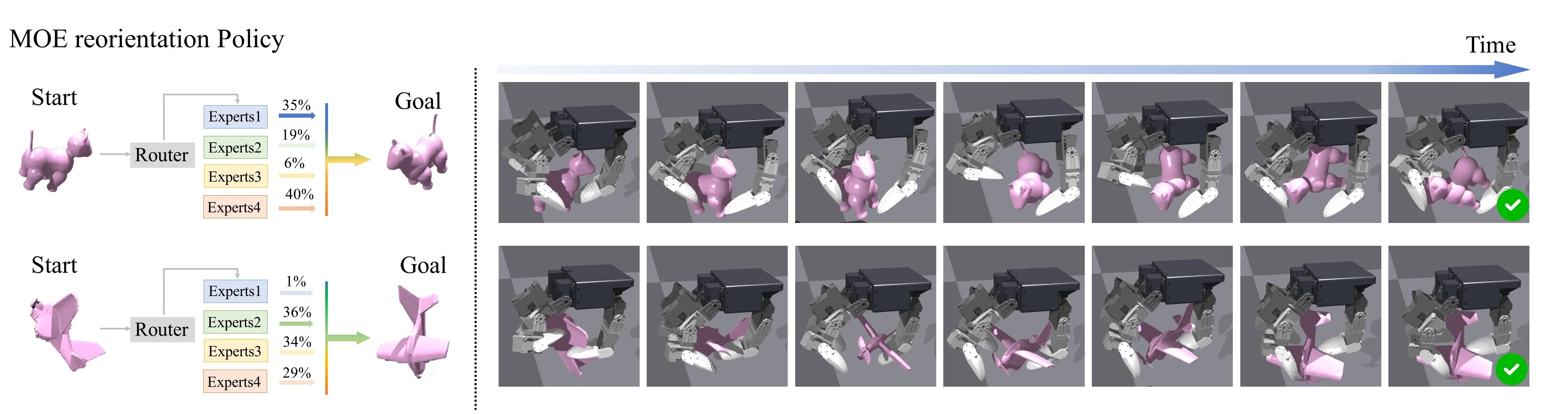}
  \caption{Visualization of DexReMoE in action. \textbf{Left:} the router adaptively allocates weights to expert policies according to the object's geometry, enabling a coordinated action generation by multiple experts.  \textbf{Right:} temporal evolution of in-hand reorientation under the same policy, showing smooth and precise rotation over time.}
  \label{fig:action1}
\end{figure*}

\begin{abstract}
In hand object reorientation provides capability for dexterous manipulation, requiring robust control policies to manage diverse object geometries, maintain stable grasps, and execute precise complex orientation trajectories. However, prior works focus on single objects or simple geometries and struggle to generalize to complex shapes. In this work, we introduce DexReMoE (Dexterous Reorientation Mixture-of-Experts), in which multiple expert policies are trained for different complex shapes and integrated within a Mixture-of-Experts (MoE) framework, making the approach capable of generalizing across a wide range of objects. Additionally, we incorporate object category information as privileged inputs to enhance shape representation. Our framework is trained in simulation using reinforcement learning (RL) and evaluated on novel out-of-distribution objects in the most challenging scenario of reorienting objects held in the air by a downward-facing hand. In terms of the average consecutive success count, DexReMoE achieves a score of 19.5 across a diverse set of 150 objects. In comparison to the baselines, it also enhances the worst-case performance, increasing it from 0.69 to 6.05. These results underscore the scalability and adaptability of the DexReMoE framework for general-purpose in-hand reorientation.

\end{abstract}

\section{Introduction}
Dexterous manipulation has advanced for a few objects~\cite{andrychowicz2020learning,handa2023dextreme,qi2023hand}, yet realizing generalizable dexterous manipulation remains a significant challenge in robotics~\cite{qi2023general}. In daily life, humans rely heavily on the remarkable versatility of their hands to perform tasks such as rearranging objects, loading dishes, tightening bolts, and slicing vegetables. Replicating this level of control in robotic systems is still extremely difficult~\cite{qi2025simple}.
At the heart of this problem lies in-hand object reorientation. A robot must be able to take an object presented in any initial pose and rotate it precisely to a desired target orientation. The ability to reliably reorient objects is crucial for flexible tool use. For example, a screwdriver must be correctly aligned with a screw before it can function properly. By focusing on this fundamental skill, we take a step closer to equipping robots with the adaptability and precision of the human hand.

Recently, the development of RL~\cite{mnih2015human,silver2017mastering,schrittwieser2020mastering} has paved the way for significant advances in dexterous manipulation research~\cite{nagabandi2020deep,andrychowicz2020learning}. In 2018, OpenAI~\cite{andrychowicz2020learning} demonstrated that a purely end-to-end deep RL pipeline could endow a multi-fingered robotic hand with unprecedented dexterity in contact-rich in-hand manipulation tasks, sparking a surge of interest despite the complexity of their sim-to-real transfer approach.. Building on this, DeXtreme~\cite{handa2023dextreme} was introduced as a vision-based system trained in Isaac Gym~\cite{makoviychuk2021isaac} with extensive domain randomization and a learned pose estimator, successfully transferring agile reorientation policies from simulation to an Allegro Hand in the real world. Meanwhile, rapid motor adaptation relying exclusively on proprioceptive history and training on simple cylindrical objects enabled a fingertip-only controller to rotate dozens of diverse real objects about the z-axis without further fine-tuning, with stable finger gaits emerging naturally~\cite{qi2023hand}. More recently, the Visual Dexterity framework, driven by a depth camera and capable of reorienting novel, complex shapes over multiple axes in real time, was presented, demonstrating generalization to unseen geometries under gravity~\cite{chen2023visual}. Despite these remarkable advances, reliably reorienting complex objects under generalized conditions remains a formidable challenge~\cite{chen2023visual}.

By leveraging transfer learning, robotic systems can generalize policies learned on a limited object set to novel scenarios with minimal additional supervision~\cite{karni1998acquisition}. The most common strategy is fine-tuning, in which pretrained parameters are adapted using only a few target-task examples. However, fine-tuning typically tailors policies to individual objects and can overwrite previously acquired skills, leading to catastrophic forgetting and limiting robustness across diverse geometries. Domain adaptation techniques~\cite{andrychowicz2020learning,handa2023dextreme} have also been investigated, but these methods focus on closing the sim-to-real gap rather than on handling substantial variation in object shape.

Another challenge is to extract meaningful object features, particularly shape information, in a computationally efficient manner. Training directly on each object's full point cloud can capture detailed geometry~\cite{chen2023visual}, but the large number of points slows learning and increases resource demands. To address these issues, we adopt a low-dimensional extrinsics embedding that encodes each object's critical properties (local surface geometry, mass distribution and pose) into a concise vector. We then extend this embedding by incorporating a point-cloud-based shape encoding together with a one-hot category vector~\cite{qi2023hand}. This representation provides the controller with a unified and expressive view of each object's physical attributes while the category information helps the router assign expert weights more effectively.

In extensive simulation experiments involving more than hundreds of complex object models, our DexReMoE surpasses monolithic baselines in consecutive success count, convergence speed, and resistance to disturbances. Moreover, it maintains these advantages when tested on objects outside the training distribution, demonstrating strong generalization and stability. These results show that a policy formed by combining multiple expert strategies and using an extrinsics embedding to encode object features can effectively tackle the challenging task of in-hand manipulation for objects with complex shapes.

In light of the above, generalizing in-hand reorientation to objects with complex shapes is still an outstanding challenge in dexterous manipulation, our work proposes a new direction for improving generalization (Figure~\ref{fig:action1}). We will release our codebase and simulation environment to facilitate further research in dexterous manipulation. In the following sections, we first review related work on object reorientation and mixture of experts methods, then describe the proposed architecture and training procedure in detail, and finally present a comprehensive experimental evaluation and analysis.

The main contributions of this paper are summarized as follows:
\begin{enumerate}
    \item We propose DexReMoE for in-hand reorientation, enabling assignment of suitable expert policies based on object geometry to accomplish the reorientation task. This framework learns a unified control policy that achieves reliable and precise in-hand repositioning.
    \item We propose a novel object shape representation that integrates point-cloud encoding with a one-hot category vector within the existing input decoupling framework. We then fuse this enhanced shape descriptor with physical properties and compress the combined features into a compact vector using a low-dimensional extrinsics embedding. This enriched representation significantly increases the expressiveness of object features.
    \item We evaluate our method on over hundreds of objects with significant shape variation, both within and outside the training distribution. Performance is measured by consecutive success counts. Extensive experimental results for comparison and ablation study demonstrate the effectiveness of the proposed method.
\end{enumerate}

\begin{figure*}[htbp]
  \centering
  \includegraphics[width=\textwidth]{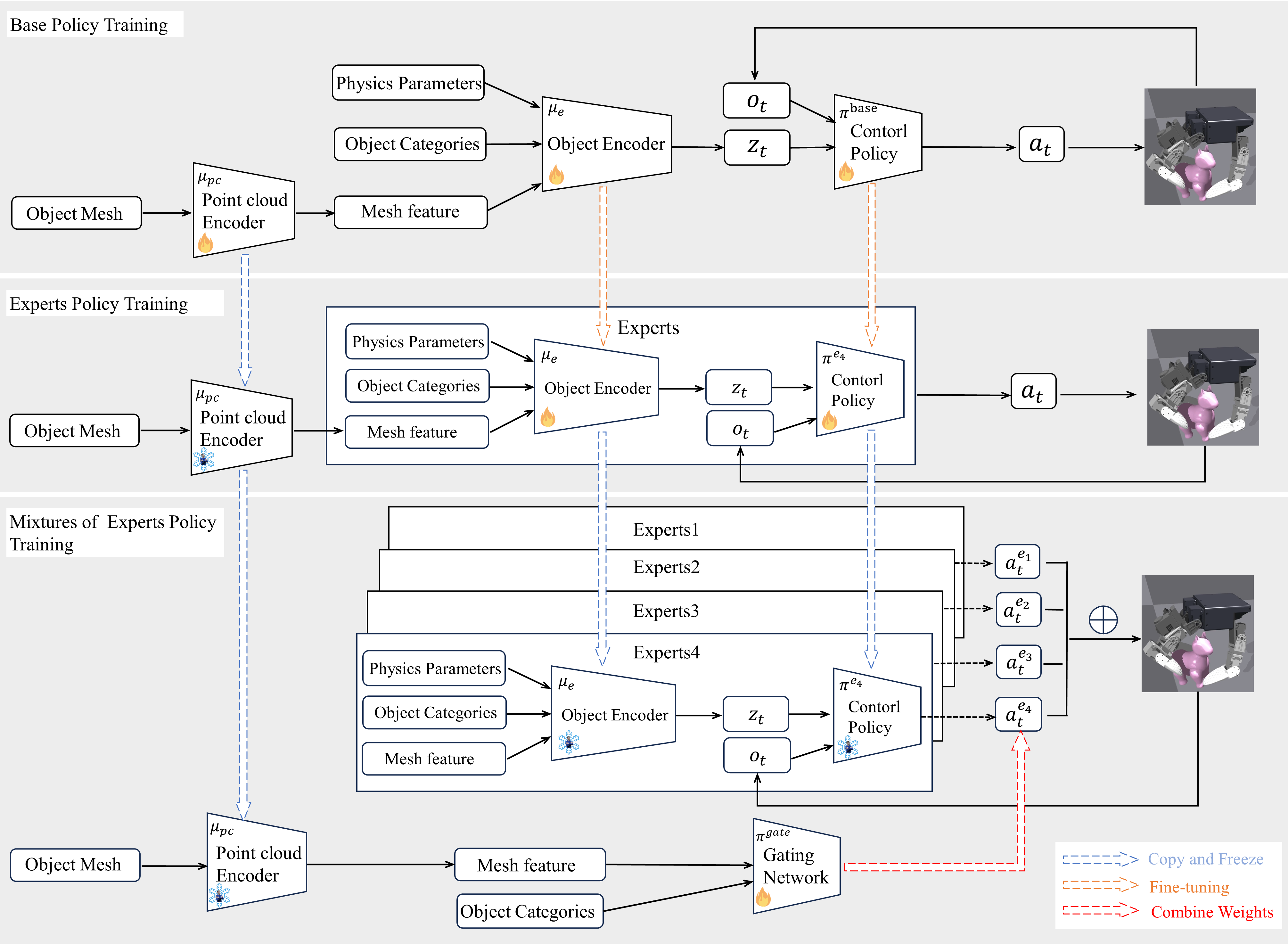} 
  \caption{An overview of our model across training. In \textbf{Base Policy Learning}, we jointly train the perception backbone $\mu_{\mathrm{pc}}$(based on PointNet++~\cite{qi2017pointnet++}), $\mu_{\mathrm{e}}$ and generalist policy $\pi_{\mathrm{base}}$, using observations $o_t$ that include the last three joint positions, commanded actions. Next, in \textbf{Experts Policy Training:} We fine-tune $\pi^{\mathrm{base}}$ to obtain four expert policies $\{\pi^{\mathrm{e_i}}\}_{i=1}^n$. Then, in \textbf{MoE Policy Training:} We freeze the  $\mu_{\mathrm{pc}}$, $\mu_{\mathrm{e}}$, and all $\pi^{\mathrm{e_i}}$. Only the soft routing network $\pi^{\mathrm{gate}}$ is trained to infer per-expert weights from the mesh feature embedding and object‐category vector, and to compute the final action via a weighted sum of the experts' outputs.}
  \label{fig:model}
\end{figure*}

\section{Related Work}
\noindent\textbf{In-Hand Dexterous Reorientation.}
In-hand dexterous reorientation has been an active research area for decades~\cite{qi2023general,saut2007dexterous,bai2014dexterous,chen2023visual,qi2025simple,andrychowicz2020learning,handa2023dextreme,sievers2022learning,openai2019solving}, with its core challenge lying in the precise coordination of finger motions to reposition, regrasp, and roll objects within constrained grasps. Early model-based methods planned stable finger trajectories using analytical representations of object and hand geometry~\cite{saut2007dexterous}, but their applicability remained limited by the complexity of real-world physics and the diversity of objects. More recently, reinforcement learning has emerged as a promising approach for complex in-hand tasks. Prior work has focused either on continuous rotations around a single axis~\cite{qi2023general,sievers2022learning}, or on multi-axis spins tailored to specific objects~\cite{qi2025simple,andrychowicz2020learning,openai2019solving}. Studies that incorporate visual inputs~\cite{chen2023visual} demonstrate that a single policy can manipulate multiple distinct objects, including those unseen during training. However, such experiments typically involve regular shapes and demand substantial training resources, so reorienting objects with complex geometries remains a significant challenge. To address this, we introduce a MoE framework consisting of a gating network and multiple expert policies, which dynamically select the most suitable expert according to object geometry to achieve reliable reorientation.

\noindent\textbf{Shape Representation in Reorientation.}
Accurate encoding of object geometry plays a critical role in any in-hand reorientation system. The dual demands of computational efficiency and policy generalization make it an open problem for dexterous hand manipulation. Prior work often sidesteps this challenge by choosing a single simple object. For example, Dextreme~\cite{handa2023dextreme} focuses exclusively on a cube and therefore requires no explicit shape encoding, but it cannot extend to other geometries. Visual Dexterity~\cite{chen2023visual} employs raw point clouds to represent shape, yet the sheer volume of points drives up computation and fails on highly symmetric objects, since point-cloud views remain unchanged under many rotations. Recently, purely tactile approaches learn a shape agent from fingertip torques and joint positions~\cite{pitz2024learning}, but it demand very high-fidelity sensors. In contrast, our method extracts compact point-cloud features via PointNet++~\cite{qi2017pointnet++}, combine them with a one-hot object categories vector, and produces a lightweight expressive embedding suitable for general dexterous reorientation.

\noindent\textbf{Mixture of Experts.}
MoE was originally introduced in \cite{jacobs1991adaptive,jordan1994hierarchical}, combining multiple specialized expert networks with a trainable gating module that adaptively weights each expert's output \cite{shazeer2017outrageously,fedus2022switch}. Recent advances in large language models have leveraged sparse MoE layers to route tokens dynamically into dedicated subnetworks, yielding both modularity and highly scalable inference \cite{xue2024openmoe,lin2024moe}. In RL, early studies demonstrated that ensembles of expert policies can capture complementary action distributions \cite{doya2002multiple,peng2019mcp}, and more recent work has leveraged MoE to advance multi-task learning in robotics, highlighting its effectiveness in coordinating diverse control objectives \cite{cheng2023multi,huang2024efficient}. In this research, we adopt the MoE framework to diversify redirection strategies in a multi-task dexterous manipulation setting. Each expert is trained on a single shape category to develop its own redirection behavior, thereby fostering broad generalization across varied object geometries.

\section{In hand reorientation with MoE}
We begin this section by outlining the overall system architecture in Section~\ref{sec:dexremoe}. Subsequently, we detail the training process for the base policy in Section~\ref{sec:basepolicy}, and conclude with the description of the MoE training procedure in Section~\ref{sec:moetrain}.
\subsection{DexReMoE}
\label{sec:dexremoe}
We propose DexReMoE, a framework that combines category specific expert fine-tuning with shared encoder representations and MoE to provide versatile and efficient in-hand reorientation across diverse object geometries. Figure~\ref{fig:model} illustrates an overview of our framework.

\noindent\textbf{Multi-Task Reinforcement Learning Framework.}  
RL formulates sequential decision making as a Markov Decision Process \(\mathcal{M}=(\mathcal{S},\mathcal{A},P,r,\gamma)\), where \(\mathcal{S}\) and \(\mathcal{A}\) denote the state and action spaces, \(P(s'\!\mid\!s,a)\) the transition probability, \(r(s,a,s')\) the immediate reward, and \(\gamma\in(0,1)\) the discount factor. A stochastic policy \(\pi_\theta(a_t\!\mid\!s_t)\), parameterized by \(\theta\), defines a distribution over actions given the current state. The learning objective is to identify parameters \(\theta^*\) that maximize the expected discounted return as:
\[
J(\theta) = \mathbb{E}_{\pi_\theta}\Bigl[\sum_{t=0}^\infty \gamma^t\,r(s_t,a_t,s_{t+1})\Bigr].
\]
In the standard single-task RL setting, an agent selects its action at time \(t\) according to:
\begin{equation*}
  a_t = \pi(s_t),
\end{equation*}
where \(s_t\) denotes the current state and \(\pi\) is the learned policy. To accommodate a diverse set of object geometries, we treat each substantially different shape category as an individual task. In conventional multi-task reinforcement learning, different tasks may have distinct objectives, reward formulations, or transition dynamics, even if they share the same state–action space. In contrast, our formulation employs a unified reward function and identical state representations across all tasks.

Accordingly, we construct our overall policy by combining the outputs of \(n\) specialized sub-policies:
\begin{equation*}
  a_t = g\bigl(\pi_1(s_t),\,\pi_2(s_t),\,\dots,\,\pi_n(s_t)\bigr),
\end{equation*}
where \(g\) denotes a generic aggregation function that determines how the individual policies are integrated into a single action. In this work, we investigate both the design of the aggregation mechanism \(g\) within a multi-task context and the selection of state representations \(s_t\) that most effectively capture inter-task shape variations.

\noindent\textbf{DexReMoE System.}  
An overview of our system architecture is provided in Figure~\ref{fig:model}. The learning process is divided into two stages: base policy training and the subsequent MoE policy training phase.

In the first stage, we jointly train a base control policy \(\pi^{\mathrm{base}}\), a point cloud encoder \(\mu_{\mathrm{pc}}\), and an object encoder \(\mu_{\mathrm{e}}\) using data collected from all object categories. The point cloud encoder extracts geometric features from raw object meshes, while the object encoder fuses these features with auxiliary information, such as object class and physical parameters, to produce a compact object representation. This representation, together with the current observation, is fed into the base policy to generate control actions.

Once training converges, we freeze both encoders and initialize a set of expert policies \(\{\pi^{\mathrm{e}_i}\}_{i=1}^4\) using the parameters of the base policy. Each expert is then fine-tuned on data restricted to a specific shape category, allowing it to specialize in manipulation behaviors tailored to a particular class of geometries. Notably, the policy inputs remain consistent across both training stages, enabling the pre-trained encoders to be reused without modification during expert specialization.

During deployment, a lightweight gating network is used to adaptively blend the outputs of the specialized experts. The gating module takes as input the object representation and a category vector, and produces a set of weights that indicate the relative importance of each expert. The final action is obtained by computing a weighted sum over the expert outputs. This modular design achieves both targeted specialization and broad generalization across diverse object shapes.

\subsection{Base Policy Training}
\label{sec:basepolicy}
\noindent\textbf{Privileged Information.}
Privileged information at time \(t\) is the concatenation of the object's physical state \(\bm{e}_t^{\mathrm{phys}}\in\mathbb{R}^{23}\) and its shape descriptor \(\bm{e}_t^{\mathrm{shape}}\in\mathbb{R}^{38}\).  The physical state is
$
  \bm{e}_t^{\mathrm{phys}}
  = [\,m,\;\bm{c},\;f,\;s,\;\bm{x}_t,\;\bm{q}_t,\;\bm{v}_t,\;\bm{\omega}_t\,],
$
where \(m\) is mass, \(\bm{c}\) center of mass, \(f\) friction coefficient, \(s\) uniform scale, \(\bm{x}_t\) position, \(\bm{q}_t\) orientation quaternion, \(\bm{v}_t\) linear velocity, and \(\bm{\omega}_t\) angular velocity.
To obtain the shape descriptor, we sample the object's point cloud and apply PointNet++~\cite{qi2017pointnet++} to extract a 100-dimensional feature \(\bm{p}_t\).  A learned point-cloud encoder \(\mu_{\mathrm{pc}}\) then maps \(\bm{p}_t\) to a 32-dimensional embedding \(\bm{f}_t\).  Appending the six-dimensional one-hot category vector \(\bm{c}\in\{0,1\}^{6}\) produces
$
  \bm{e}_t^{\mathrm{shape}} = [\,\bm{f}_t,\;\bm{c}\,].
$
Concatenating \(\bm{e}_t^{\mathrm{phys}}\) and \(\bm{e}_t^{\mathrm{shape}}\) yields the full privileged vector
$
  \bm{e}_t = [\,\bm{e}_t^{\mathrm{phys}},\;\bm{e}_t^{\mathrm{shape}}\,],
$
which is passed through an encoder \(\mu_{\mathrm{e}}\) to produce the 66-dimensional embedding
$
  \bm{z}_t = \mu_{\mathrm{e}}(\bm{e}_t).
$
The resulting embedding 
$
  \bm{z}_t = \mu_{\mathrm{e}}(\bm{e}_t)
$
serves as the policy's privileged input.  We refer to \(\bm{z}_t\) as the extrinsics embedding and observe that it significantly enhances generalization across diverse objects and environments.

\noindent\textbf{Observations and Outputs.}
In our formulation, we define the policy input state as the combination of the robot's proprioceptive observation \(\bm{o}_t\) and a privileged object encoding \(\bm{z}_t\). This composite representation captures both the robot's recent behavior and essential object-specific information. The base policy \(\pi^{\mathrm{base}}\) receives this full state and produces an action \(\bm{a}_t\) to be executed by the PD controller. Specifically, the observation \(\bm{o}_t\) encodes a short temporal window of joint positions and previously applied actions:$
  \mathbf{o}_t = \bigl[\bm{q}_{t-2},\,\bm{q}_{t-1},\,\bm{q}_t,\,\bm{a}_{t-3},\,\bm{a}_{t-2},\,\bm{a}_{t-1}\bigr],
$
where each \(\bm{q}_t\) denotes the joint positions at time \(t\), and \(\bm{a}_{t-k}\) refers to the executed action at time \(t-k\). The policy output is then given by:
$
  \mathbf{a}_t = \pi^{base}\bigl(\mathbf{o}_t,\;\mathbf{z}_t\bigr).
$
To improve the smoothness of control, we apply exponential moving average to the action outputs rather than using them directly:
$
\overline{\bm{a}_t} = \alpha\,\bm{a}_t + (1 - \alpha)\,\overline{\bm{a}_{t-1}}
$
where \(\alpha \in [0, 1]\) is a smoothing coefficient. Empirically, we observe that smaller values of \(\alpha\) lead to increased training difficulty due to diminished action responsiveness.

\noindent\textbf{Reward Function.}
Our reward function (Eq.~\eqref{eq:regularization_reward}) is composed of several components. The first term in the reward function represents the task's success criterion; within a fixed time horizon, each successful placement of the object at the target location yields a reward, and accumulating multiple successes encourages the dexterous hand to perform consecutive repositioning operations. However, this success-based reward alone is sparse and provides limited learning signals, making it insufficient for stable policy learning. To address that, we incorporate additional reward shaping terms to guide the learning process. Specifically, we penalize the relative distance \(|\delta_p|\) and orientation difference \(|\delta_\theta|\) between the object and the target pose to encourage the agent to minimize both positional and rotational discrepancies . Moreover, we introduce a penalty on action magnitude and joint velocities to further promote smoother control. The reward terms are mathematically expressed as:

\begin{equation}
\left\{
\begin{aligned}
    & r_{1} = c_{\text{success}} \\ 
    & r_{2} = c_{\text{dist}} \cdot |\delta_p| + \frac{c_{\text{rot}}}{|\delta_\theta| + \epsilon} \\ 
    & r_{3} = c_{\omega} \sum_{i=1}^{n} \left[ |\omega_{i,t}| - \omega_{\text{clip}} \right]_+ + c_{a} \cdot \|a_t\|_2^2,
\end{aligned}
\right.
\label{eq:regularization_reward}
\end{equation}
where \(c_{\text{success}}>0\) is the reward for reaching the target; \(c_{\mathrm{dist}}, c_{\mathrm{rot}}, c_{\omega}, c_{a}<0\) are penalty weights; \(\epsilon\) prevents division by zero; \(n\) is the hand's degrees of freedom; \(\omega_{i,t}\) is the angular velocity of joint \(i\) at time \(t\); \(\omega_{\mathrm{clip}}\) is the velocity threshold; and \(a_t\) is the action vector at time \(t\).  

The total reward at each timestep is then defined as: 
\begin{equation}
R = r_{1} + r_{2} + r_{3}.
\end{equation}

Our reward function does not include the penalty for the object falling, as we found during experiments that such a term suppresses exploratory actions and adversely affects the overall training performance. The proposed reward function enables direct training of reorientation policies capable of operating in air.

\noindent\textbf{Policy Optimization.}
We employ Proximal Policy Optimization~\cite{schulman2017proximal} to simultaneously train both the policy $\pi^{\mathrm{base}}$ , the embedding module $\mu_{\mathrm{e}}$ and $\mu_{\mathrm{pc}}$ . The weights between the policy and the critic network are shared, with an extra linear projection layer to estimate the value function. During training, each environment is initialized with an object in a random pose. Since our training directly targets in-air object manipulation, we initialize the dexterous hand with a stable grasp configuration to ensure faster convergence during training.

\subsection{MoE Training}
\label{sec:moetrain}
To accommodate the full spectrum of object geometries, from perfectly flat surfaces and slender elongated forms to highly intricate topologies, we enhance our base reorientation policy with a MoE architecture. Rather than relying on a single network to cover all shape variations, we introduce multiple specialist sub-networks, each dedicated to capturing specific geometric features. the gating network assigns each expert a continuous weight based on the object's geometry, and the final control action is obtained by computing the weighted average of all expert outputs.

\noindent\textbf{Expert Knowledge.} 
Each expert policy is realized as an independent neural network with its own parameter set, which enables specialization in distinct regions of the shape‐action manifold. By decoupling representation learning across experts, we avoid forcing a monolithic policy to cover all geometric variations simultaneously. The total number of experts \(n\) is determined by the intrinsic correlations among object classes rather than by the count of objects, thereby preventing model complexity from scaling linearly with dataset size. In our implementation, four experts share a frozen point-cloud encoder \(\mu_{\mathrm{pc}}\) to preserve a common perceptual backbone while each expert refines only its private policy head. The generalist expert is fine-tuned on a broad set of object shapes to provide baseline reorientation capabilities; the airplane expert specializes in handling elongated, discontinuous surfaces; the train expert focuses on slender structures with high aspect ratios; and the complex animal expert is designed to manage non-uniform surfaces and intricate topologies. This modular framework supports efficient extension or pruning as new shape categories emerge without retraining the entire ensemble.

\noindent\textbf{Task-Specific Feature Extraction and Soft Router Formulation.}  
We designate the object's geometry as the sole task-specific feature, encoded by a compact shape descriptor \(\bm{e}_t^{\mathrm{shape}}\). This choice ensures that the extracted features capture the most salient, discriminative aspects of each object while remaining invariant throughout the reorientation process, thereby simplifying policy optimization and improving convergence. Concretely, \(\bm{e}_t^{\mathrm{shape}}\) is obtained by fusing a point-cloud encoding with a one-hot category vector and compressing the result via a low-dimensional extrinsics embedding.  

The gating network \(\pi^{\mathrm{gate}}\) maps the shape descriptor \(\bm{e}_t^{\mathrm{shape}}\) to a vector of scores, these scores are normalized via softmax to yield nonnegative weights, which we then use to perform a weighted aggregation of the experts' outputs. We demonstrate in the following chapter through experiments that this dense soft gating converges much more reliably than hard Top-K routing, since it avoids abrupt switches between experts and better accommodates sparse reward signals. To clarify the underlying mathematical structure of this mechanism, we introduce the notation and steps of the algorithm. Formally, let \(\bm{e}_t^{\mathrm{shape}}\in\mathbb{R}^d\) denote the shared descriptor vector and \(\{f_i\}_{i=1}^n\) be our \(n\) expert mappings \(f_i:\mathbb{R}^d\to\mathbb{R}^h\). The gating network \(\pi^{\mathrm{gate}}\) is implemented as a two-layer MLP with learnable weight matrices \(W_1\in\mathbb{R}^{64\times d}\) and \(W_2\in\mathbb{R}^{n\times 64}\), using the ELU~\cite{clevert2015fast} activation function between layers. The Soft Mixture-of-Experts algorithm then proceeds as follows:
\begin{enumerate}
  \item \textbf{Gating network.} Compute unnormalized expert scores
  \[
    \ell_i = \bigl[W_2\,\mathrm{ELU}(W_1\,\bm{e}_t^{\mathrm{shape}})\bigr]_i,
    \quad i = 1, \dots, n,
  \]
  and normalize to obtain routing weights
  \[
    p_i = \frac{\exp(\ell_i)}{\sum_{j=1}^n \exp(\ell_j)}.
  \]

  \item \textbf{Expert evaluations.} Each expert produces
  \[
    \bm{y}_i = f_i(\mathbf{x})
  \]

  \item \textbf{Soft aggregation.} The final output is
  \[
    \bm{y}
    = \sum_{i=1}^n p_i\,\bm{y}_i.
  \]
\end{enumerate}

\section{Experiment}
We first describe the experimental setup in Section~\ref{sec:experiment}. In Section~\ref{sec:baseline}, we present our evaluation metrics and baseline methods for quantitative comparison.
Next, in Section~\ref{sec:base-policy}, we introduce the simulator-state-based base policy and evaluate how object geometry affects its performance. In particular, we analyze how variations in shape complexity influence the policy's convergence and robustness across a wide range of objects.
In Section~\ref{sec:ablation}, we report ablation studies on various design choices. 
Then, we examine the limitations of the base policy when dealing with objects of complex geometry and present our proposed MOE framework, emphasizing its performance benefits in Section~\ref{sec:moe}. Finally, Section~\ref{sec:tsen} uses t-SNE to cluster the gating network's weight vectors, revealing distinct regions for varied shape complexities and coherent groupings for similar geometries.

\subsection{Experiment Setup}
\label{sec:experiment}

\noindent\textbf{Simulation Setup.}
We employ the IsaacGym simulator \cite{makoviychuk2021isaac} to train our skill policy, planner policy and state estimator. Both simulation and control frequencies are set to 60 Hz. During training, we run 32768 parallel environments to collect samples for agent training. Related approaches typically perform object reorientation on a tabletop or with the palm oriented upward. In contrast, our configuration executes fully mid-air reorientation with the palm directed downward. This arrangement is significantly more complex and prone to failure, making the task more challenging. In our experiments, we adhere to the standard protocol by reorienting objects entirely in mid-air with target orientations randomly sampled from the SO(3) space using our custom GX11 three-fingered dexterous hand \cite{dong2025gex}, a manipulator with eleven degrees of freedom.We train on a dataset of 150 object models sourced from online repositories such as Google Scanned Objects \cite{downs2022google}. These models span a broad spectrum of intricate nonconvex geometries, including vehicles, footwear and various animal forms, ensuring comprehensive coverage of complex shape categories.

\noindent\textbf{Reorientation Success Criterion.}
We quantify manipulation performance by counting the number of consecutive successful reorientations achieved within each fixed time window. A straightforward criterion that declares success whenever the orientation error falls below a specified tolerance can be misled by incidental collisions that briefly align the object with its target pose. To eliminate these false positives, we require both precise orientation and a sustained halt at the desired configuration.  
At each control step, the reach criterion is considered satisfied only when the rotational distance to the goal is less than or equal to $\tau_\theta$, each finger joint velocity remains below $\tau_q$, the object's linear velocity stays under $\tau_v$, and its angular velocity does not exceed $\tau_\omega$.

To guard against transient alignment, we enforce that all four conditions hold continuously throughout the final control cycle of the episode. A reorientation is deemed successful only when this sustained-hold criterion is met, ensuring the policy learns precise alignment and stable maintenance rather than relying on incidental collisions. This capability is essential in real-world tasks where the robot must maintain a tool's pose for subsequent actions.

\subsection{Evaluation Metrics and Baseline Methods.}
\label{sec:baseline}
To evaluate the performance of the proposed DexReMoE algorithm, we employ five summary metrics, denoted as follows:

\begin{equation}
\left\{
\begin{aligned}
S_{\min}     &= \min_{i} S_i \\
S_{\max}     &= \max_{i} S_i \\
\bar S_{5-}  &= \frac{1}{5}\sum_{i\in\mathcal{W}_5} S_i \\
\bar S_{5+}  &= \frac{1}{5}\sum_{i\in\mathcal{B}_5} S_i \\
\bar S       &= \frac{1}{N}\sum_{i=1}^{N} S_i,
\end{aligned}
\right.
\label{eq:metrics}
\end{equation}
where, \(S_i\) denotes the consecutive success count for object \(i\), \(\mathcal{W}_5\) and \(\mathcal{B}_5\) are the index sets of the five lowest- and highest-performing objects respectively, and \(N\) is the total number of test objects. Table~\ref{tab:baseline_compare} reports these five metrics for all evaluated methods.  

To verify the superiority of our method, we use the following RL algorithms as baselines: 1) the ideal policy trained with Domain Randomization (DR) by OpenAI~\cite{andrychowicz2020learning}; 2) the Privileged Feature (PrivFeat) policy~\cite{qi2023hand}, which uses privileged physical state information in place of raw observations; 3) the Privileged Shape (PrivShape) policy~\cite{qi2023general}, which incorporates additional object shape information (e.g., point clouds) as privileged input; 4) the Adaptive Domain Randomization (ADR) policy~\cite{handa2023dextreme}, which adjusts domain parameters based on task performance; 5) the Residual Actions (Res) policy~\cite{ceola2024resprect}, which learns a residual correction on top of a pre-trained base policy for high-dimensional dexterous hand control ;6) the Sparse Mixture-of-Experts (SparseMoE) policy~\cite{shazeer2017outrageously}, which activates a sparse subset of experts via a  noisy gating network; 7) the Switch Transformer (Switch) policy~\cite{fedus2022switch}, which routes each input to a single expert using a learned switching mechanism; 8) the Low-Rank Expert Mixture (MLoRE) policy~\cite{yang2024multi}, which augments the MoE architecture with a shared convolutional path for task-invariant feature extraction, and adopts low-rank convolutional experts to reduce parameter and computational overhead; and 9) the Multi-gate Mixture-of-Experts (MMoE) policy~\cite{ma2018modeling}, which shares expert sub-networks across tasks and employs task-specific gates to combine experts and model inter-task relationships. Both the baseline are trained with the same reward and penalty settings as our method.

\begin{figure}[t]    
  \centering
  \includegraphics[width=\columnwidth]{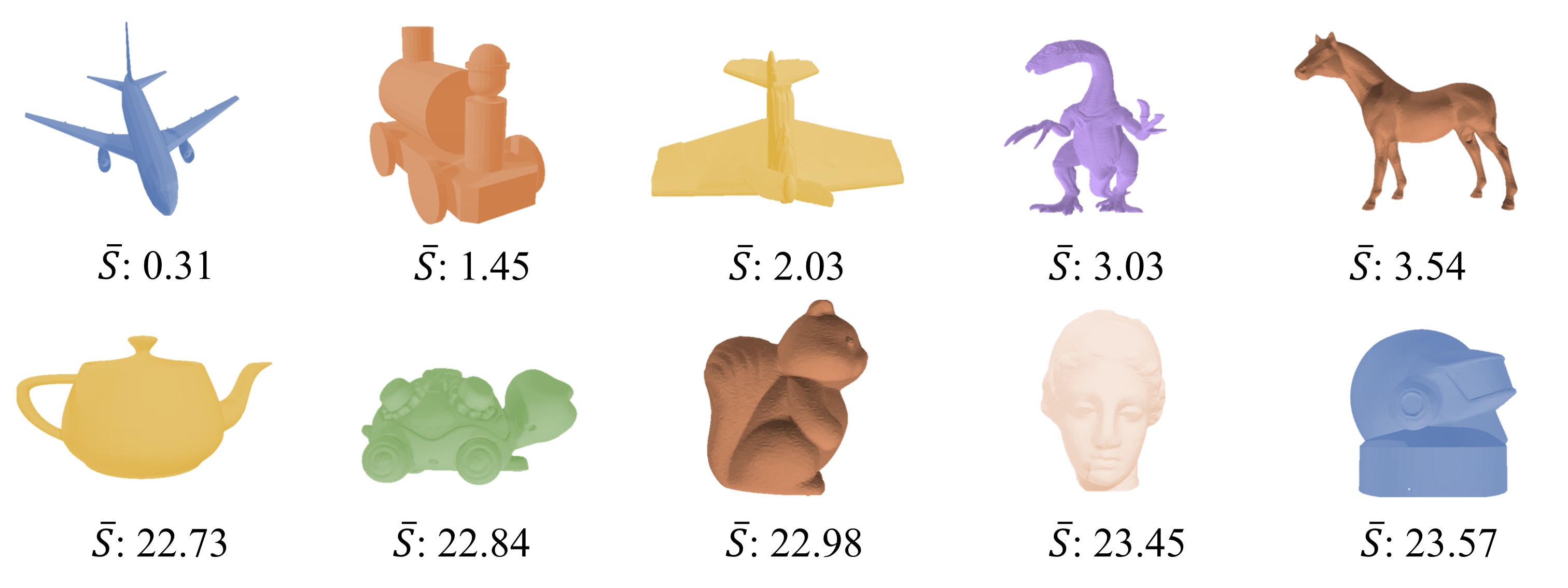}
  \caption{Top: the five worst-performing objects with their consecutive success counts $\bar S$. Bottom: the five best-performing objects.}
  \label{fig:low_best_object}
\end{figure}

\subsection{Impact of Object Geometry on Base Policy Performance.}
\label{sec:base-policy}
We tested the base policy on 100 objects from the training set in 6000 episodes. Empirically, we observe that the majority of objects can be successfully reoriented more than 15 consecutive times, with some achieving up to 23. However, a small subset of objects consistently fails to succeed even once. To better understand how object geometry affects reorientation performance, we first identify the five best-performing and five worst-performing objects under the base policy. Their shapes, along with their corresponding mean consecutive success counts \(\bar S\), are visualized in Figure~\ref{fig:low_best_object}. We then analyze the reorientation trajectories of the low-performing group and observe that failures frequently arise when protruding components become lodged between the fingers, preventing further rotation until the episode times out. For instance, repeated jamming occurs with airplane models, where wing tips obstruct motion, and with train model, whose elongated chassis often becomes trapped during manipulation. These outcomes suggest that objects with pronounced protrusions or extreme aspect ratios present substantial learning difficulties due to their increased shape complexity. 

\begin{figure*}[!t]
  \centering
  \includegraphics[width=\textwidth]{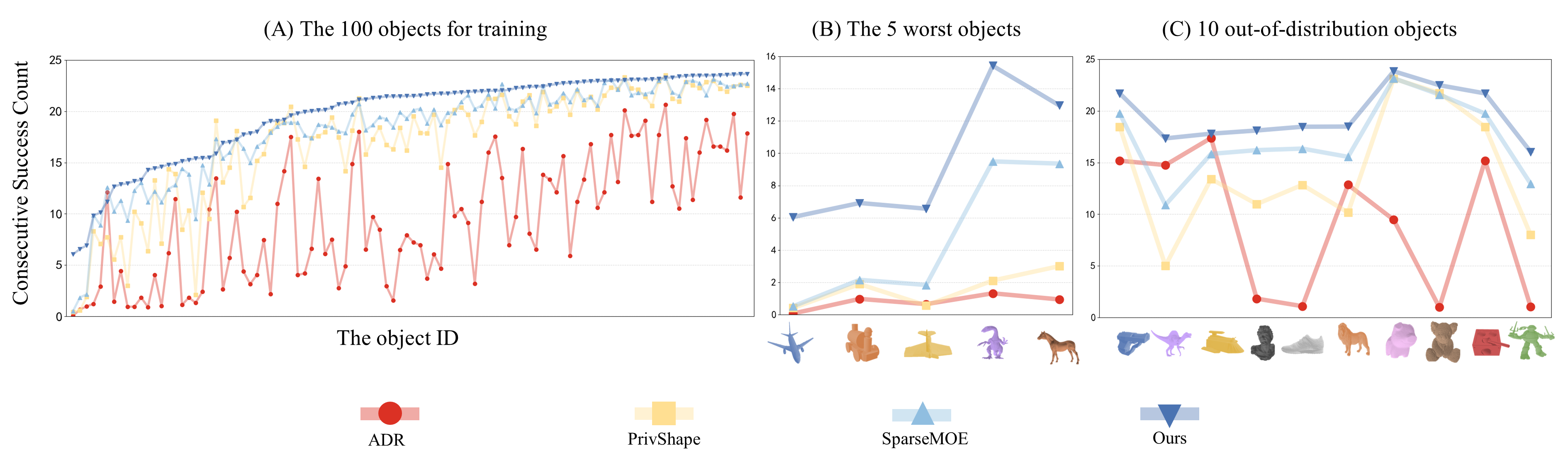} 
\caption{{Comparison of Consecutive Success Count Across Baselines}. (a) compares the consecutive success count of our policy against the baseline across all 100 objects, with objects ordered from lowest to highest success under our method. It is clear that our approach consistently outperforms the baseline on the majority of items. (b) focuses on the five objects that the baseline struggled with the most. While the baseline's performance remains poor on these challenging shapes, our policy achieves substantially higher and more reliable success rates. (c) presents results for ten objects randomly selected from outside the training domain. For simpler shapes, both methods achieve similar levels of consecutive successes; however, as object complexity increases, our policy continues to maintain a clear advantage over both baseline strategies.}
\label{fig:baseline}
\end{figure*}

\begin{figure}[!t]
  \centering
  \includegraphics[
    width=\columnwidth,       
    keepaspectratio,
    clip
  ]{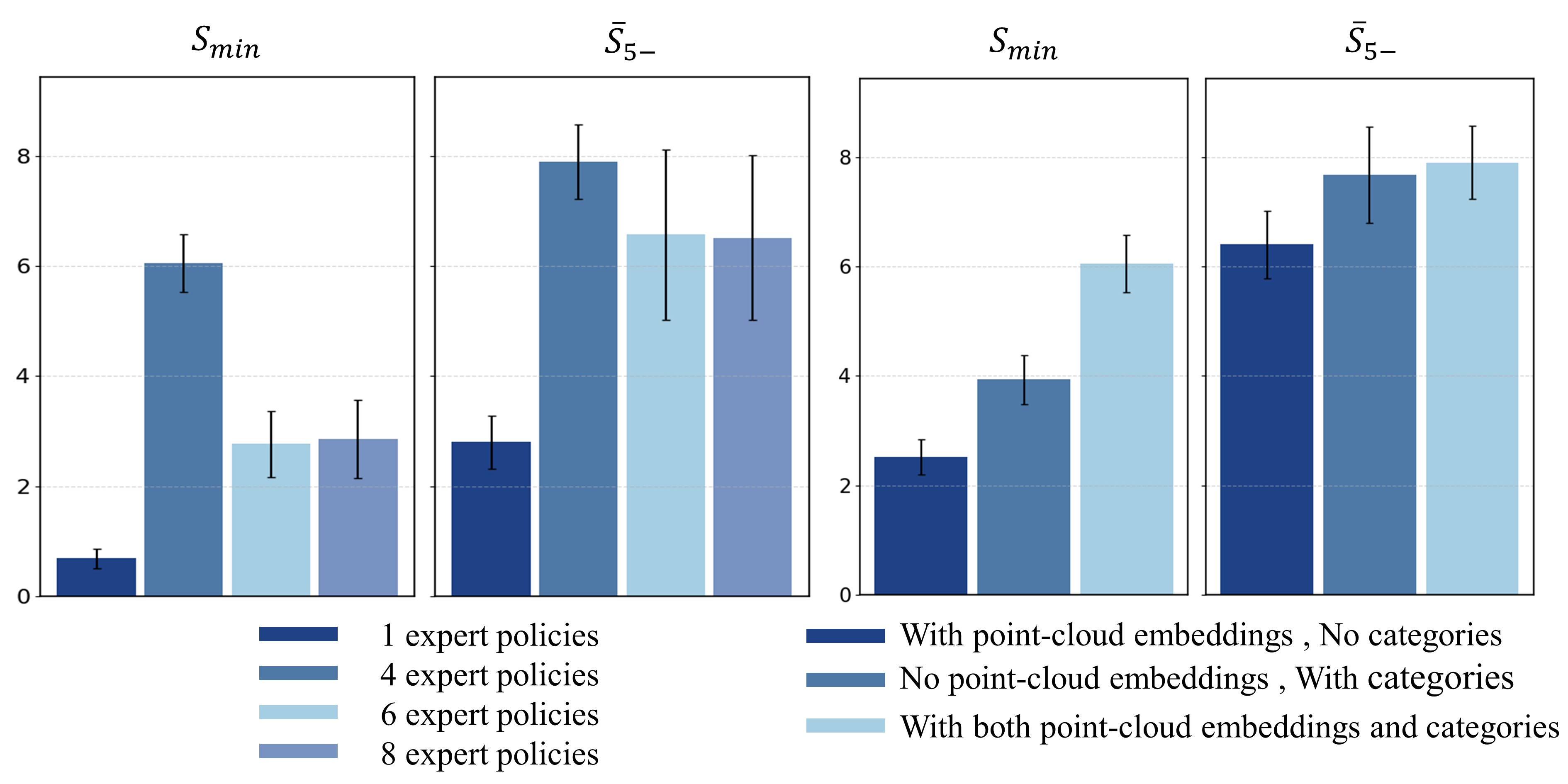}
\caption{Ablation Experiments. \textbf{Left}: Performance of MoE policies with varying numbers of expert policies. \textbf{Right}: Impact of different inputs to the gating network. Error bars indicate the standard deviation of the performance metric computed over 6,000 episodes.}
  \label{fig:ablation}
\end{figure}

\subsection{Ablation Experiments}
\label{sec:ablation}
In addition to the design of Soft MoE policy, we also make several critical design choices within our architecture. In this section, we examined two key design choices: the number of Experts and the inputs provided to the gating network. All experiments were conducted on the same set of 100 objects used during training to ensure a fair comparison across conditions.

\noindent\textbf{Number of Expert Policies.}
We investigated how the number of expert policies affects performance in our Soft MoE framework by comparing configurations with 1, 4, 6, and 8 expert policies. The single-expert case corresponds to a conventional base policy without any specialization. In the four-expert setup, three specialists were trained on the poorest-performing object classes while a single generalist policy addressed all others. The six-expert configuration allocates one expert per object category. Finally, the eight-expert configuration adopts a finer-grained taxonomy to further partition object types, yielding eight distinct experts. As illustrated in Figure \ref{fig:ablation}(Left), the four-expert configuration unexpectedly achieves superior $S_{\min}$ and $\bar S_{5-}$ compared with both outperforming both the single-expert and the larger expert variants. We hypothesize that exceeding an optimal expert count degrades performance because routing inefficiencies and diminished per-expert data lead to imbalanced training and overfitting, which together undermine both specialization and generalization on the most challenging geometries. These findings highlight that, beyond a certain point, more experts do not necessarily translate to better performance; rather, a suitable expert count offers the best trade-off between expressivity and robustness.

\noindent\textbf{Gating Network Inputs.}
To assess the impact of gating network inputs on reorientation performance, we compared the full router (which ingests both point-cloud embeddings and category embeddings) against two ablated variants: one driven solely by the point-cloud embedding and another relying exclusively on the category embedding. As show in Figure~\ref{fig:ablation} (Right), although the two configurations achieved similar average success counts, the model with the additional categories embedding produced noticeably better results on the worst single object and on the worst five objects. The observed improvement implies that category embeddings provide global directional cues to counteract performance drops on complex shapes, with point cloud data simultaneously refining expert probability distributions.

\renewcommand{\arraystretch}{1.2}
\begin{table*}[!t]
\centering
\renewcommand{\arraystretch}{1.3}
\begin{tabular}{lccccc|ccccc}
\hline
\multirow{2}{*}{Method}
  & \multicolumn{5}{c|}{Within Training Distribution}
  & \multicolumn{5}{c}{Out‐of‐Distribution} \\
\cline{2-6}\cline{7-11}
& $S_{\min}^\uparrow$ & $S_{\max}^\uparrow$ & $\bar S_{5-}^\uparrow$ & $\bar S_{5+}^\uparrow$ & $\bar S^\uparrow$
& $S_{\min}^\uparrow$ & $S_{\max}^\uparrow$ & $\bar S_{5-}^\uparrow$ & $\bar S_{5+}^\uparrow$ & $\bar S^\uparrow$ \\
\hline
DR~\cite{andrychowicz2020learning}        
  & 0.11 & 23.52 & 0.84  & 22.15 & 11.38
  & 0.09 & 21.42 & 1.20  & 20.79 & 11.59 \\

PrivFeat~\cite{qi2023hand}
  & 0.31 & 23.48 & 2.06  & 22.74 & 15.13
  & 1.71 & 23.51 & 3.60  & 23.29 & 16.59 \\

PrivShape~\cite{qi2023general} 
  & 0.41 & 23.5 & 1.59 & 23.12  & 16.93
  & 2.59 & 23.42 & 3.97 & 23.21  & 16.25 \\

ADR~\cite{handa2023dextreme}       
  & 0.14 & 23.53 & 0.64 & 23.1  & 12.32
  & 0.85 & 23.52 & 1.79 & 23.13  & 12.44 \\

Res~\cite{ceola2024resprect}       
  & 0.70 & 23.52 & 2.09  & 23.29 & 15.62
  & 0.53 & 23.33 & 2.48  & 21.01 & 13.00 \\

SparseMoE~\cite{shazeer2017outrageously} 
  & 0.52 & 23.50 & 4.68  & 23.43 & 19.02
  & 3.03 & 23.47 & 7.50  & 23.26 & 17.45 \\

Switch~\cite{fedus2022switch}    
  & 0.66 & 23.52 & 2.67  & 23.33 & 18.33
  & 2.27 & 23.59 & 5.39  & 23.31 & 16.85   \\

MLoRE~\cite{yang2024multi}     
  & 1.49 & 23.24 & 4.88  & 22.91 & 17.35
  & 2.71 & 23.24 & 5.99  & 23.07 & 16.64 \\

MMoE~\cite{ma2018modeling}      
  & 3.36 & 23.35 & 7.42  & 23.17 & 18.97
  & 3.80 & 23.49 & 8.67  & 23.35 & 18.18 \\

Ours      
  & \textbf{6.05} & \textbf{23.56} & \textbf{7.90} & \textbf{23.43} & \textbf{19.62}
  & \textbf{4.11} & \textbf{23.69} & \textbf{9.14} & \textbf{23.53} & \textbf{19.12} \\
\hline
\end{tabular}
\caption{We compare our method to several baselines in simulation under two evaluation settings: (1) \textit{Within Training Distribution}; (2) \textit{Out-of-Distribution}. Each method is evaluated across five metrics: the minimum consecutive success count $S_{\min}$, the maximum $S_{\max}$, the average of the five worst-performing objects $\bar{S}_{5-}$, the average of the five best-performing objects $\bar{S}_{5+}$, and the overall average $\bar{S}$. Our approach performs exceptionally well on objects with complex surface geometries in both the training-distribution and out-of-distribution scenarios, significantly outperforming the baseline methods. 
}
\label{tab:baseline_compare}
\end{table*}

\subsection{Evaluation of the Soft MoE Policy}
\label{sec:moe}

\noindent\textbf{Soft MoE Policy Performance.}
We first assess all methods on objects drawn from the same distribution used during policy training. To provide quantitative comparisons, Table \ref{tab:baseline_compare} summarizes five key metrics: minimum and maximum single-object consecutive successes ($S_{\min}$, $S_{\max}$), mean performance on the five hardest and easiest objects ($\bar S_{5-}$, $\bar S_{5+}$), and the overall average ($\bar S$).
Among the baselines, MoE-based schemes (e.g., MMoE and SparseMoE) deliver respectable overall means near 19 but still suffer from low worst-case performance ($S_{\min}$ below 1 for SparseMoE). Traditional methods such as ADR and PrivShape lag further behind, with overall averages under 17 and minimal robustness on the most challenging shapes.

In contrast, our Soft MoE policy dramatically elevates the performance floor by raising \(S_{\min}\) from under 1 to 6.05, and it matches or exceeds ceiling performance, achieving \(\bar S_{5+}=23.43\) and \(S_{\max}=23.56\). Overall, it achieves \(\bar S = 19.62\) consecutive successes, outperforming all nine baselines while demonstrating both stronger worst-case guarantees and consistently high success across the entire object set.

Figure~\ref{fig:baseline}(A) plots the consecutive-success counts for each of the 100 objects, sorted by performance under our model. For clarity, we compare only three representative baselines (ADR, PrivShape and SparseMoE) to highlight the relative gains. Our policy surpasses each of these methods on the vast majority of shapes, demonstrating its ability to generalize across varied surface geometries. Figure~\ref{fig:baseline}(B) then focuses on the five most challenging objects for these three baselines, which were previously unsolvable under their policies. With our approach, those objects become reliably reorientable with multiple consecutive successes in every trial.

When initially reproducing the baseline DR and ADR~\cite{andrychowicz2020learning,handa2023dextreme} under our strict success criterion, which requires maintaining a stable hold at the goal orientation, training failed to converge. Specifically, the monolithic policy never achieved reliable in-hand rotations across the full spectrum of complex geometries. To restore performance, we implemented curriculum learning. This process began with a relaxed success test using a single cube, then progressively tightened the criterion while incrementally introducing all 100 objects. Only after this staged progression did the baseline achieve comparable results in our evaluation.
Conversely, our approach utilizes a modular reinforcement learning framework that explicitly segregates object-centric inputs from hand-centric state-action information. This decoupled architecture not only facilitates more effective feature learning but also eliminates the necessity for curriculum training. From the initial epoch onward, the policy acquires robust reorientation behaviors within a single training phase. 

\noindent\textbf{Out-of-Distribution Robustness.}
We then study the out-of-distribution robustness of object shapes for a trained model. We begin by assessing each policy's ability to handle object shapes that lie beyond the training distribution. Figure~\ref{fig:baseline}(C) shows consecutive-success counts on ten randomly selected out-of-distribution objects. For the simple objects, both our method and the baseline achieve comparable success count. However, as surface complexity increases, the baseline's performance degrades sharply, ultimately failing entirely on some items, whereas our policy continues to deliver consistent, high-success count.

As shown in Table~\ref{tab:baseline_compare}, the strongest baseline, MMoE, already exhibits zero-shot capability. It achieves an out-of-distribution mean of 18.18 consecutive successes. However, it still collapses on the most challenging shapes, registering a minimum success count of only 3.80 and an average of 8.67 on the five hardest objects. By contrast, our method raises the minimum consecutive-success count on these difficult items to 4.11 and boosts the mean across the five most challenging objects to 9.14.

These findings confirm that our method not only excels on familiar objects but also preserves its advantage when confronted with novel, complex geometries, achieving strong zero-shot performance without any further training or adaptation.

\subsection{Clustering Analysis of Gating Network Outputs}
\label{sec:tsen}
\begin{figure}[!htb]  
  \centering
  \includegraphics[width=\columnwidth]{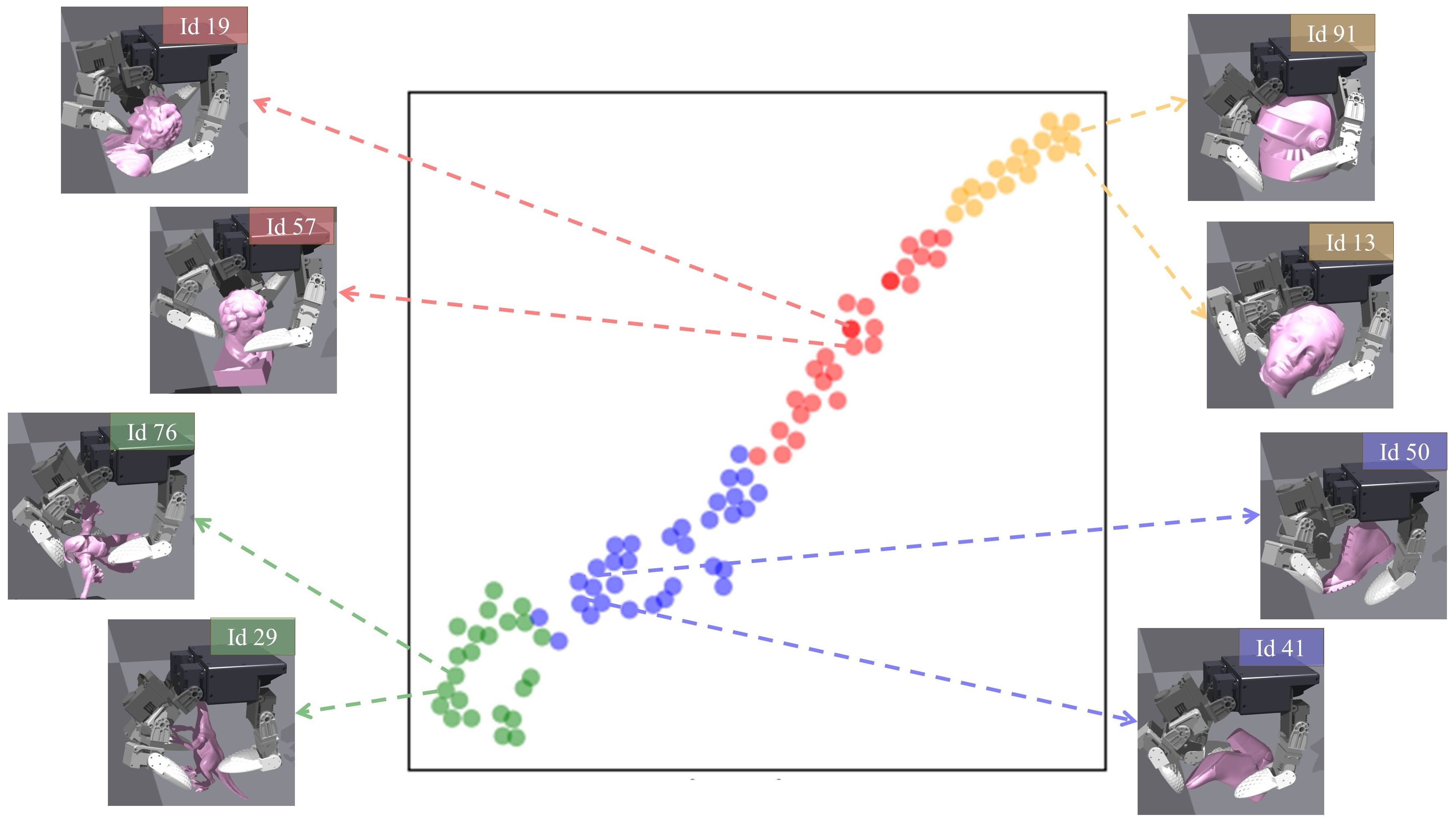}
  \caption{t-SNE projection of gating-network weight vectors for 100 objects. Points are colored by cluster assignment, showing that objects with different geometries occupy distinct regions while similar shapes form the same clusters.}
  \label{fig:tsen}
\end{figure}

To understand how the gating network $\pi^{\mathrm{gate}}$ differentiates object geometries when assigning experts, we employed t-SNE (Figure~\ref{fig:tsen}) to visualize the expert assignment weight vectors produced by the gating network $\pi^{\mathrm{gate}}$ after it was trained on 100 object models with rotational augmentation for 6000 times. We found that objects of different geometric complexity, as reflected by their consecutive success rates under the base policy, occupy separate regions in the embedding space. For example, objects that achieve high consecutive success rates tend to cluster in the upper-right quadrant, such as object IDs 91 and 93, while those with lower rates group in the lower-left quadrant. Moreover, objects with similar shapes form tighter clusters, for instance IDs 41 and 50 (both shoes) and IDs 19 and 57 (both sculpted human heads). These observations indicate that the gating network effectively captures geometric distinctions and assigns experts in a shape-dependent manner. 

\section{Conclusion and limitations}
In this work, we introduced the Soft MoE policy for in-hand object reorientation and demonstrated its successful deployment across a variety of complex shapes. By leveraging multiple specialized experts, our approach efficiently adapts to differing object geometries during training. Experiments show that the Soft MoE architecture not only achieves reliable performance on known objects but also generalizes effectively to new, unseen shapes without additional retraining.
Our results underscore the potential of Mixture-of-Experts networks in robotic policy learning. Specifically, incorporating diverse expert models enhances training efficiency, bolsters robustness against challenging object geometries, and improves overall generalization. These findings suggest that Soft MoE frameworks can serve as a powerful tool for developing adaptable, high-performing robotic manipulation strategies.

\noindent\textbf{Limitations and Future Work:}
Our reliance on manually labeled object categories limits scalability. To overcome this, we will explore multimodal large-language models to automate labeling. Moreover, we have yet to validate Soft MoE on physical hardware; conducting real-world trials is a primary goal for our next research phase.
\bibliographystyle{IEEEtran}  

\newpage
\appendix
\section*{Experimental Details}
\begin{figure}[!h]    
  \centering
  \includegraphics[width=\columnwidth]{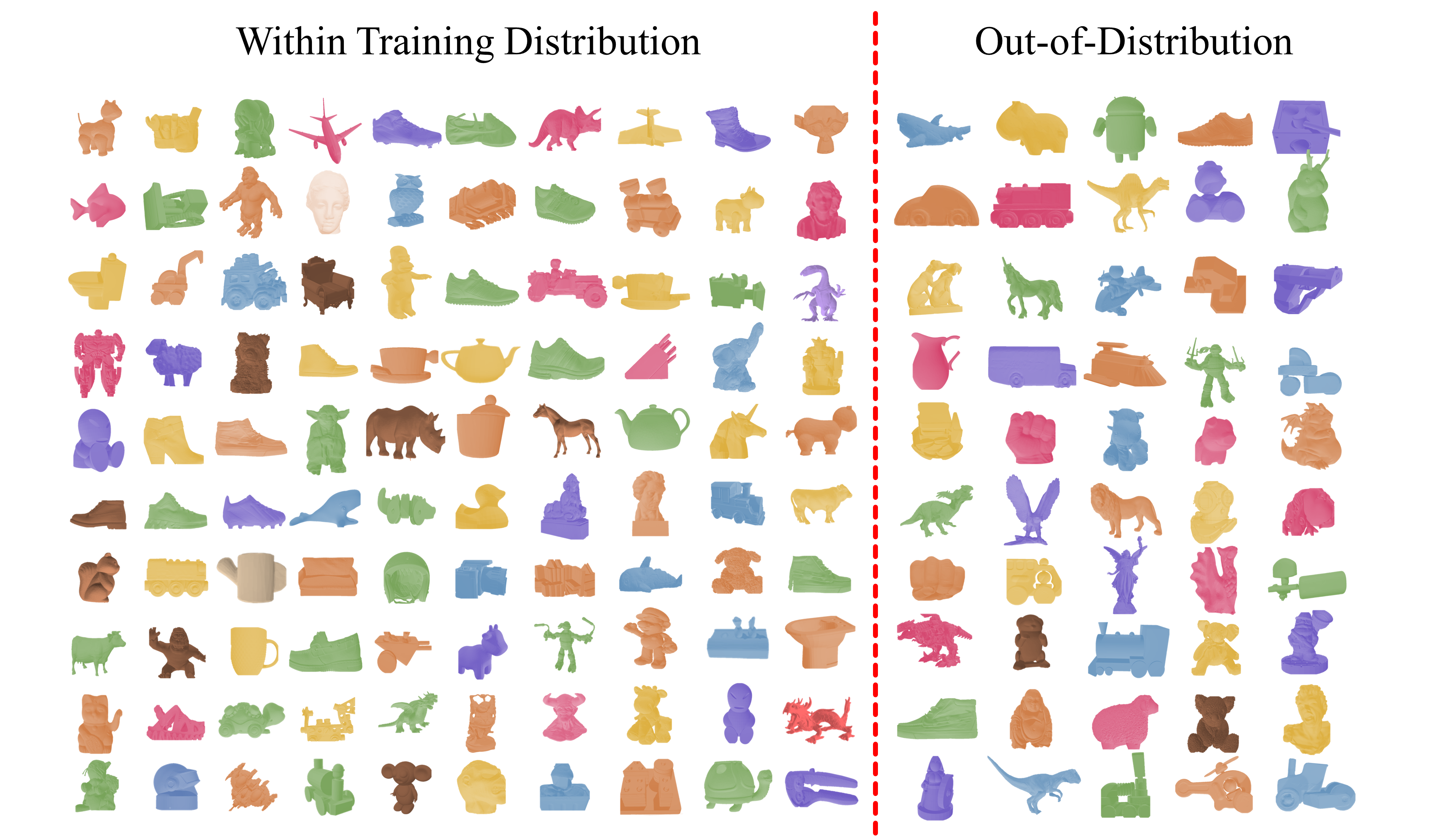}
  \caption{Overview of the complete object dataset, on the left of the red line, we show the training dataset. And on the right of the red line, we show the out-of-distribution (testing) dataset.}
  \label{fig:dataset}
\end{figure}

\noindent\textbf{Object Dataset:}
We employ the full set of 150 objects from our dataset (see Figure~\ref{fig:dataset}), selecting 100 at random for training and reserving the remaining 50 for out-of-distribution evaluation. To ensure that each mesh can be manipulated by the robotic hand, we first center it and then scale it by a factor of 0.8 so that its dimensions align with the hand's workspace in simulation. We observed that scaling meshes below a certain threshold, such as reducing them to 60\% of their original size, shifts manipulation from precise fingertip control to collisions with the inner surfaces of the fingers. Moreover, when an object becomes very small, its complex geometric features no longer convey meaningful distinctions, and the dexterous hand effectively treats it as a simple, diminutive cube.

\begin{figure}[!h]     
  \centering
  \includegraphics[width=\columnwidth]{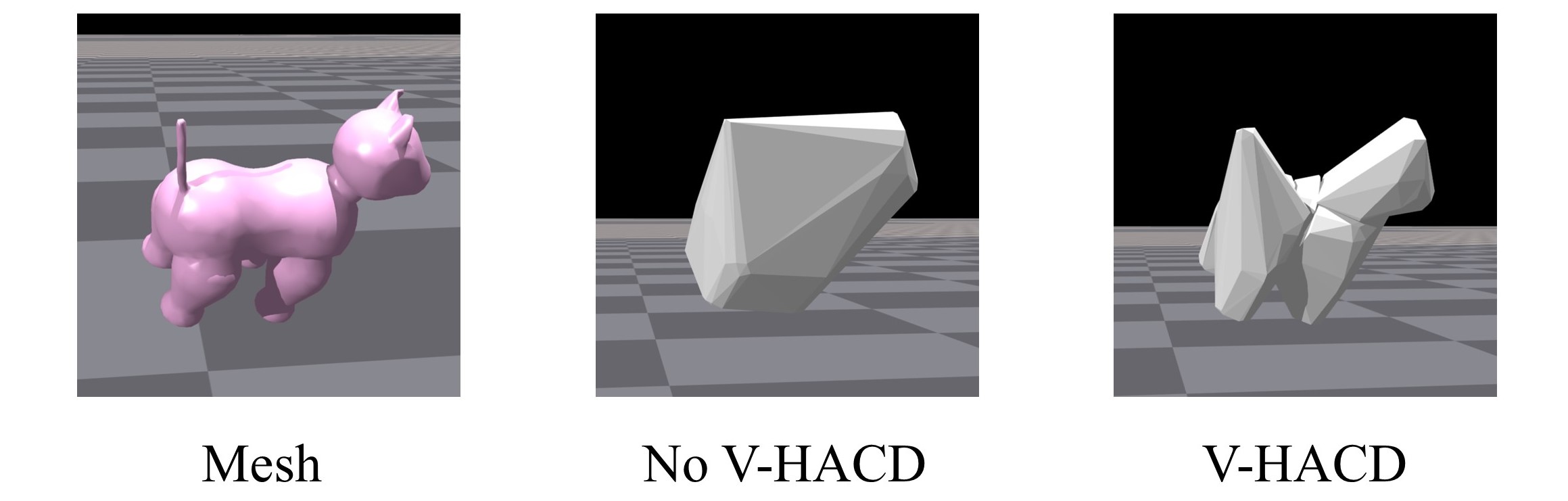}
  \caption{We show the difference between object meshes with and without convex decomposition.}
  \label{fig:vhacd}
\end{figure}

\noindent\textbf{Convex Decomposition:}
We use approximate convex decomposition (V-HACD~\cite{mamou2016volumetric}) to perform an approximate convex decomposition on the object and the robot hand meshes for fast collision detection in the simulator (Figure~\ref{fig:vhacd}). 

\noindent\textbf{Policy architecture:}
All networks are implemented as MLPs and trained with Adams~\cite{kingma2014adam}: the base policy $\pi^{\mathrm{base}}$ uses two hidden layers of 512 units each; the point-cloud encoder \(\mu_{\mathrm{pc}}\) has three layers of 32 units; the object encoder \(\mu_{\mathrm{e}}\) comprises two layers of 256 and 128 units; and the gating network $\pi^{\mathrm{gate}}$ consists of two 64-unit layers with ELU activations~\cite{clevert2015fast}.

\noindent\textbf{Hyper-parameters:}
Table~\ref{tab:hyperparams} lists the hyper-parameters used in the experiments.

\vspace{-2pt}

\begin{table}[!h]
  \centering
  \caption{Hyper-parameter Setup}
  \label{tab:hyperparams}
  \begin{tabular}{llll}
    \toprule
    Hyper-parameter         & Value    & Hyper-parameter        & Value     \\ 
    \midrule
    num of envs            & 32768    & episode length          & 600       \\
    horizon length         & 8        & minibatch size         & 16384     \\
    learning rate           & 5e-3     & PPO clip range         & 0.2          \\
    kl threshold            & 0.02     & PPO gamma         & 0.99       \\
    PPO tau            & 0.95           &success tolerance      & 0.4       \\
    $c_{\text{success}}$    & 800      & $c_{\mathrm{dist}}$    & -10.0     \\
    $c_{\mathrm{rot}}$      & -1.0      & $c_{a}$                & -0.0002   \\
    $\tau_{\theta}$         & 0.1      & $\tau_{q}$             & 10.0      \\
    $\tau_{v}$              & 0.04     & $\tau_{\omega}$        & 0.5       \\
    \bottomrule
  \end{tabular}
\end{table}

\end{document}